# ClinEdge: An Expert-Graded Benchmark of Real Patient Edge Cases Where Frontier Multimodal Models Fail

**Dr. Anurag Sharma[1] Aditya Sharma[2] Sapna Baalwan[3] for the ClinEdge Clinical Review Panel[4]**

[1] Orthopaedic Surgeon; [2] FutrBridge; [3] Independent Researcher; [4] Panel members are listed under Authors, Ethics, and Disclosures.

Correspondence: research@futrbridge.com | https://www.futrbridge.com/research

## Abstract

Most benchmarks for large language models in medicine are built from licensing examinations and theoretical questions. The few that draw on real clinical data do so only in part, and rarely require a model to read a patient's own imaging. ClinEdge addresses this gap with 32 real, de-identified patient cases from active clinical practice, each evaluated across seven frontier systems for a total of 224 expert-graded evaluations. The cases pair primary imaging with short vignettes; each contains a feature known to defeat pattern-matched reasoning, and each is scored against the patient's documented clinical course. The suite is built for depth rather than volume. Grading followed a two-stage expert process. First, a senior orthopaedic surgeon gave all 224 responses a pass-or-fail verdict and also scored every free-response answer on a weighted four-part rubric. Second, a panel of nineteen physicians, largely senior specialists at teaching medical colleges, independently reviewed every case and every one of the 224 grades. All seven production systems were evaluated through their public interfaces, and all 224 conversations are preserved as replayable links.

Performance was poor across all systems, and the failures were shared: different vendors missed the same findings and converged on the same wrong plans. From the experts' notes we catalogue 18 recurring failure modes in five families. The models stop reading an image at the first finding. They give generic plans and lean toward over-treatment. They trust a misleading history over correct imaging. Their answers vary across systems and runs, and change with question format. Failing answers were delivered fluently and without any expressed uncertainty. Four cases are released in full; the remaining 28 are held out, and we invite frontier AI labs and healthcare AI companies to evaluate their models against them (research@futrbridge.com).

## 1 Introduction

Most of the evidence for clinical ability in language models comes from examinations and other theory-based tests. Models pass licensing question banks [1], perform strongly on knowledge benchmarks [2, 3], and do well on conversational health rubrics [4]. Those results measure theoretical understanding. The work a clinician does on a hard case looks different. It means reading an image against an imperfect history, deciding what the image actually shows, including what a hurried reader would miss, and then committing to one plan for one patient, knowing that a plausible-sounding wrong plan is more dangerous than an honest deferral.

ClinEdge measures that gap directly. The cases are real, de-identified patient encounters from active clinical practice, selected by experienced medical professionals using one criterion: each case had to contain a specific feature they expect to defeat pattern matching. The ground truth for every case is the documented clinical course, not a textbook answer key, and the cases are anchored to primary imaging the system has to actually read; the vignette alone does not determine the answer, and in one control it points the wrong way on purpose. The experts validated every step of every response, not only the final answer: the reading of the imaging, the use of the case details, and the reasoning that leads to the plan. We evaluated seven systems through their public consumer interfaces at the highest reasoning settings, because that is how patients, families, and junior clinicians reach these models in practice, and every conversation was saved as a shareable link, so all 224 evaluations can be replayed and checked by anyone.

The results are poor across the board. The best system passed 8 of 32 cases; eleven cases were failed by all seven systems, often for the same reason, with everyone missing the same finding in the same film; and on 9 of the 11 multiple-choice cases a majority of systems chose the same wrong option, so a second opinion from a different vendor is rarely an independent check. No system's average quality reached 2.7 out of 5, and every system averaged below the midpoint on every dimension. Because the failures repeat across frontier-lab assistants, a big-tech assistant, open-weight models, and a search-based answer engine, they look like properties of the shared training distribution rather than any one company's mistakes.

This paper contributes:

1. **An expert-curated benchmark of real patient edge cases with a held-out design.** 32 cases centred on orthopaedics and radiology, and also spanning neurology and neurosurgery, rheumatology, internal medicine, paediatrics, ophthalmology, trauma surgery, sports injury, and hand surgery. Four cases are released in full (Appendix A); 28 are held out so the benchmark remains a clean test for future systems.
2. **An expert grading framework.** A four-dimension weighted rubric plus an independent pass or fail verdict, applied by experienced medical professionals with panel review of every case and grade, a written note behind every grade, and a replayable conversation link behind every evaluation.
3. **A rigorous evaluation of seven production systems,** reported with confidence intervals and paired significance tests rather than bare percentages.
4. **A catalogue of 18 recurring failure modes in five families** (Table 5), drawn from the experts' per-response grading notes. These are named, testable categories that model developers and benchmark builders can work against.

## 2 Related Work

Examination benchmarks such as MedQA [1], MultiMedQA [2], early GPT-4 results [3], and the harder specialty-board suite MedXpertQA [5] established that models can retrieve and apply medical knowledge at licensing-exam level, but all of them stay inside the exam paradigm: curated questions, option lists, answer-match scoring, and follow-up work shows that high multiple-choice accuracy partly reflects cues in the options, since performance drops when familiar options are changed or removed [6]. Rubric and conversational evaluations (HealthBench [4], HealthBench Professional [7], MedHELM [8]) grade open-ended responses against physician-written criteria and compare models with physician baselines [9, 10], but the decisive evidence in those tasks arrives as text. Record-grounded and agentic benchmarks (MIMIC-CDM [11], CliniCARE-Bench [12], AgentClinic [13]) evaluate how models investigate records and simulated encounters, and show that a correct final answer can hide a defective process; imaging sits at the edge of those designs. Medical visual question answering and generalist models [14, 15] score finding descriptions on curated images, and recent diagnostic studies of consumer systems on musculoskeletal radiographs report fracture-detection sensitivity far below clinicians [16, 17], matching the perception failures we document. ClinEdge sits in the space these lines of work leave open (Table 1): real patient primary imaging that must be read to clinical depth, ground truth taken from what actually happened to the patient, grading by experienced clinicians, and validation of the full reasoning process rather than the final answer alone.

**Table 1. How ClinEdge relates to existing clinical evaluations.** ✓ full, Partly = partial, — absent, based on each benchmark's own description. "Outcome-based ground truth" means answers come from what actually happened to the patient rather than an examination key.

| Benchmark | Substrate | Real patient data | Primary imaging central | Outcome-based ground truth | Graded by expert clinicians | Reasoning process validated |
|---|---|---|---|---|---|---|
| MedQA [1] | Exams and mixed QA | — | — | — | — | — |
| MultiMedQA [2] | Exams and mixed QA | — | — | — | Partly | Partly |
| MedXpertQA [5] | Specialty board exams | — | Partly | — | Partly | — |
| HealthBench [4] | Conversations and tasks | — | — | — | Partly | — |
| MedHELM [8] | Conversations and tasks | Partly | — | Partly | Partly | — |

| Benchmark | Substrate | Real patient data | Primary imaging central | Outcome-based ground truth | Graded by expert clinicians | Reasoning process validated |
|---|---|---|---|---|---|---|
| MIMIC-CDM [11] | EHR decision-making | ✓ | — | ✓ | Partly | Partly |
| CliniCARE-Bench [12] | Longitudinal EHR audit | ✓ | Partly | Partly | Partly | ✓ |
| AgentClinic [13] | Simulated encounters | Partly | Partly | — | Partly | Partly |
| **ClinEdge (this work)** | Real patient cases with imaging | ✓ | ✓ | ✓ | ✓ | ✓ |

# 3 The ClinEdge Benchmark

## 3.1 Cases, curation, and composition

All 32 cases come from real, de-identified encounters in the clinical team's practice and referral networks. For each case the ground truth is what actually happened: the imaging as reported, the operative findings where surgery took place, the implant actually in the patient, and the management that resolved the presentation. A case entered the suite only if it contained at least one feature the clinical experts, from years of practice, expect to defeat pattern matching: a hidden finding that is easy and dangerous to miss; hardware whose identification decides the plan; a bilateral condition needing grading on each side; a history that conflicts with the imaging; anatomy that rules out the default procedure; or a presentation where holding back is correct and intervening is the trap. The suite has 11 multiple-choice cases, each with one correct option among three to five plausible alternatives (several distractors are exactly the answers systems volunteered in free-response form), and 21 free-response cases. Imaging: 25 X-ray, 3 MRI, 2 X-ray with CT, 1 clinical photograph, 1 text-only. Orthopaedics and radiology are involved in nearly every case, and the suite also covers implantology (8 cases), neurology and neurosurgery (3), paediatrics (2), trauma surgery (2), and one case each in rheumatology, internal medicine, ophthalmology, sports injury, and hand surgery. Anatomically, the cases span the spine, hip and pelvis, knee, foot and ankle, and hand. Six scenarios appear twice, once in each question format, with the same patient and imaging (Section 4.3). Two cases are controls on clinical context: Case 1 pairs imaging with a history written to conflict with it, and passes only a system that reports the imaging on its own merits; Case 6 has no image at all, describing an alarming-sounding but benign presentation whose explanation sits in the vignette, and passes only a system that does not escalate. In total the benchmark yields 224 evaluations.

## 3.2 Release policy: four cases public, twenty-eight held out

ClinEdge is meant to stay useful as a test, so this paper does not disclose the full case set. Cases 8, 9, 14, and 18 are released in full in Appendix A: vignette, imaging, options and correct answer for the multiple-choice case, per-system outcomes and scores. The other 28 cases are held out. Throughout the paper, their failure analyses are written at pattern level, without revealing the diagnosis, the decisive finding, or the option set, because a case whose answer is public measures memorisation the next time it is used. A domain-level roster of all 32 cases ships with the release, and organisations can evaluate against the held-out set by contacting the authors (Section 7).

## 3.3 Systems and protocol

We evaluated seven production systems in four groups: three assistants from frontier AI labs (ChatGPT 5.6 Sol, Gemini 3.7 Flash, and Claude Fable 5 at high effort); one big-tech integrated assistant (Microsoft Copilot, Think Deeper); two open-weight models (DeepSeek Vision and Sarvam 105B); and one search-augmented answer engine (Perplexity Sonar 2.0). We used the public consumer interfaces at the highest available reasoning setting rather than APIs, on purpose. These interfaces are how patients and junior clinicians actually reach the models, and each one bundles the vision processing and settings the vendor ships as its best public product. Results therefore describe the deployed system, not the isolated base model. Each case was a fresh, single-turn conversation: the vignette exactly as written, with the imaging attached, no extra framing, and no follow-up hints. Every conversation was preserved as a shareable link.

## 3.4 Grading

**Multiple-choice cases** are graded in two steps, not by option match alone. The response must select the designated correct option, and the working behind it must also hold up: the grading expert reviewed how each system analysed the case, read the imaging and reasoned toward its choice, and a response that picked the correct option through a wrong reading or flawed methodology was marked as failed. Choosing the right option is therefore necessary but not sufficient to pass. **Free-response cases** receive two separate judgments. The first is a holistic pass or fail verdict: would this answer, acted on as given, be acceptable care for this patient? The second is a quality score on four dimensions (Table 2), each rated 1 to 5 and combined with weights that reflect clinical priority. Accuracy carries 0.40 because everything downstream depends on it; clinical reasoning carries 0.30; thoroughness and clinical relevance carry 0.15 each. The ratings cover the whole response, not the final answer alone: the interpretation of the uploaded imaging (X-ray, CT, MRI, or clinical photographs), the use of the case details, and whatever the case asks for, whether a diagnosis, a treatment plan, or ongoing management.

**Who graded, and how the grades were checked.** Grading followed the two-stage panel process that is standard for reference standards in medical AI, where independent expert review with majority-based adjudication gives more reproducible ground truth than any single reader [18]. Case curation and primary grading were led by Dr. Anurag Sharma, an orthopaedic surgeon with more than ten years of clinical practice (MS - Orthopaedics, SMS Medical College & Hospital), working against each patient's documented course and writing a note naming the specific error behind every failing grade. Every case and grade was then independently reviewed by the ClinEdge Clinical Review Panel, nineteen experienced physicians, most of them senior orthopaedic specialists, including heads of department, heads of unit, and professors at teaching medical colleges, who checked each case for clinical validity before inclusion and each grade against the saved model conversation; disagreements were resolved by discussion and majority view, and the grades reported here are the panel-agreed versions. Panel members are listed in the back matter. Every grade, note, and conversation link is kept in a single grading log covering all 224 evaluations, so each grade remains auditable against its saved conversation.

**Table 2. The weighted quality rubric.** Each dimension is rated on a five-point scale: 5 Excellent (fully correct and complete), 4 Very good (correct, with minor gaps), 3 Adequate (broadly accurate but incomplete), 2 Poor (significant errors), 1 Very poor (fundamentally incorrect). The scale applies to the whole response: the imaging, the case details, and the requested output.

| Dimension | Weight | What it measures |
|---|---|---|
| Accuracy | 0.40 | Correctness of the interpretation of the imaging and case details and of the diagnosis, including secondary findings and hardware |
| Clinical reasoning | 0.30 | The logic from findings to plan: differentials, indications, and the choice between surgery and conservative care for this patient |
| Thoroughness | 0.15 | All injuries addressed; both sides graded where the condition is bilateral; work-up and follow-up specified |
| Clinical relevance | 0.15 | Specific and actionable: the procedure and implant named, patient-specific factors respected |

**Statistics and reporting.** The case set is small and deliberately hard, so we treat uncertainty as a first-class part of reporting. Every proportion carries a 95% Wilson score interval; paired comparisons between systems use exact McNemar tests; and we read overlapping intervals conservatively, never claiming a ranking the tests cannot support. Reporting follows the applicable items of the TRIPOD-LLM guideline for studies of large language models in healthcare [19]: system identity, access route, settings, prompts, grading procedure, grader identity, and full per-evaluation provenance.

# 4 Results

## 4.1 No system passes more than one case in four

Of the 224 evaluations, 47 passed: 21.0% overall (95% CI 16.2 to 26.8%). The seven systems sit in a narrow band from 15.6% to 25.0% (Table 3, Figure 1). Copilot and Sarvam 105B lead with 8 of 32; ChatGPT 5.6 Sol and DeepSeek Vision trail with 5 of 32; the other three sit at 7 of 32. The confidence intervals overlap heavily, and even the widest comparison, best system against worst, is not statistically separable (exact McNemar test on the 11 discordant cases, $p = 0.55$). The supported finding is the band, not the ordering. On material a postgraduate surgical trainee is expected to handle, no frontier system reached 30%. The multiple-choice subset adds a note of its own: six of seven systems scored exactly the same on it (2 of 11; Perplexity 3 of 11), so on cases this hard that format barely distinguishes the systems; the separation in Table 3 comes from the free-response cases and the quality rubric.

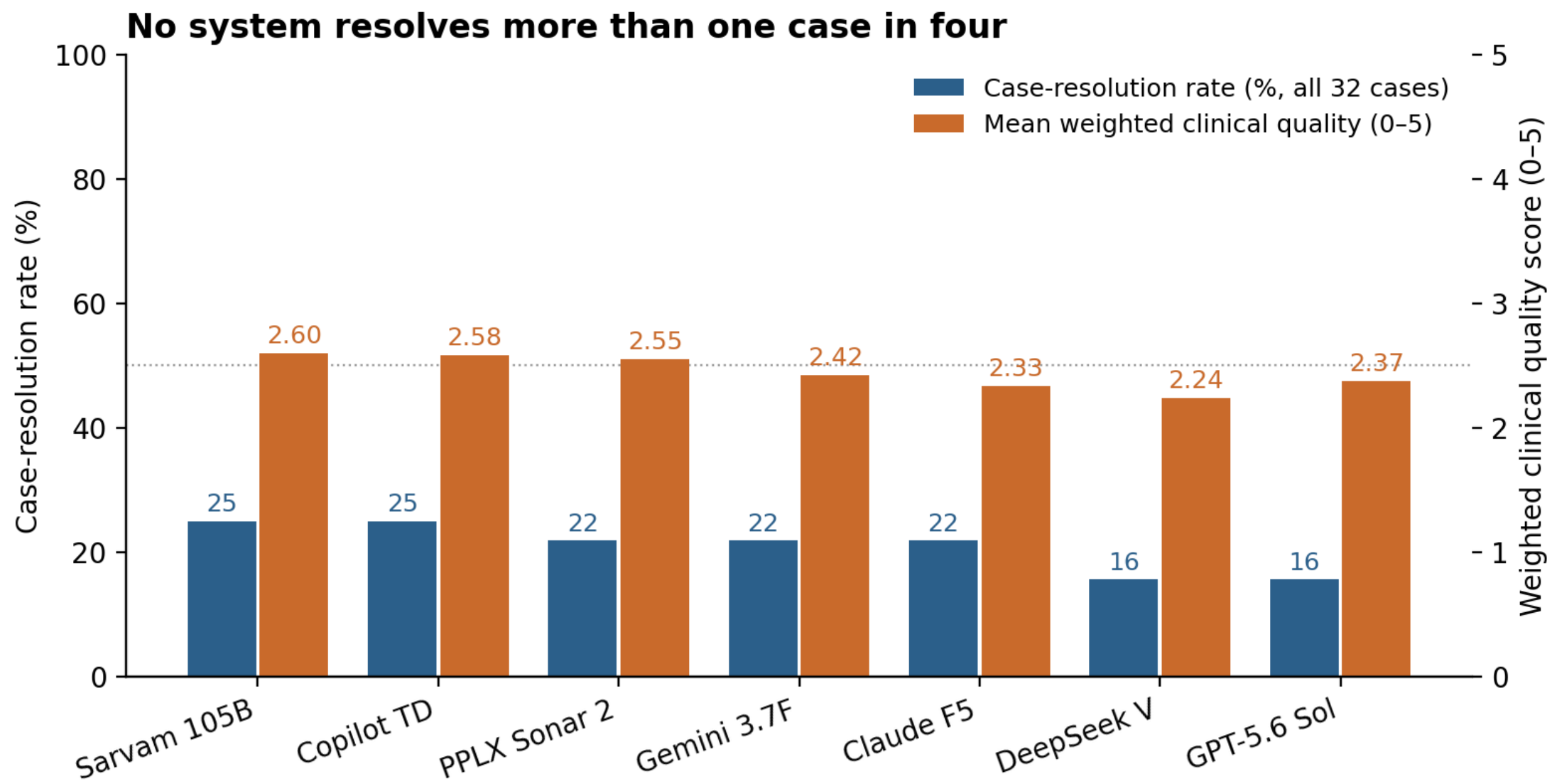

**Figure 1.** Pass rate over all 32 cases (blue, left axis) and mean expert-graded quality score on free-response cases (orange, right axis) for the seven systems.

**Table 3. Primary results.** 95% CI = Wilson score interval on the overall rate. Weighted quality is the mean expert-graded score (0 to 5) over graded free-response runs (Perplexity n = 17; its remaining runs could not be completed due to technical failure). The last four columns are per-dimension means.

| System | MCQ (of 11) | Free-resp. (of 21) | Overall (of 32) | 95% CI, overall | Weighted quality (0–5) | Acc. | Reas. | Thor. | Relev. |
|---|---|---|---|---|---|---|---|---|---|
| Copilot (Think Deeper) | 2 (18.2%) | 6 (28.6%) | **8 (25.0%)** | 13–42% | 2.58 | 2.94 | 2.88 | 2.76 | 2.88 |
| Sarvam 105B | 2 (18.2%) | 6 (28.6%) | **8 (25.0%)** | 13–42% | **2.60** | 2.94 | 2.83 | 2.72 | 2.83 |
| Gemini 3.7 Flash | 2 (18.2%) | 5 (23.8%) | 7 (21.9%) | 11–39% | 2.42 | 2.32 | 2.42 | 2.42 | 2.47 |
| Claude Fable 5 | 2 (18.2%) | 5 (23.8%) | 7 (21.9%) | 11–39% | 2.33 | 2.39 | 2.50 | 2.56 | 2.50 |
| Perplexity Sonar 2 | 3 (27.3%) | 4 (19.0%) | 7 (21.9%) | 11–39% | 2.55 | 2.69 | 2.62 | 2.56 | 2.69 |
| ChatGPT 5.6 Sol | 2 (18.2%) | 3 (14.3%) | 5 (15.6%) | 7–32% | 2.37 | 2.05 | 2.45 | 2.50 | 2.60 |
| DeepSeek Vision | 2 (18.2%) | 3 (14.3%) | 5 (15.6%) | 7–32% | 2.24 | 2.69 | 2.56 | 2.38 | 2.56 |

## 4.2 Expert-rated quality is low, and quality dissociates from verdicts

Over the 143 numerically graded free-response runs, the mean weighted quality score is 2.44 of 5. Forty-four percent of graded responses score at or below 2.0, which in the grading notes means a wrong reading compounded by wrong management, while only 18% reach 4.0 or above. Figure 2 shows the per-dimension profile: every system averages below the scale midpoint of 3 on every dimension. The weakest dimension varies by system, but accuracy, the dimension that gates everything downstream and carries the largest weight, is the weakest for three of the four lowest-quality systems.

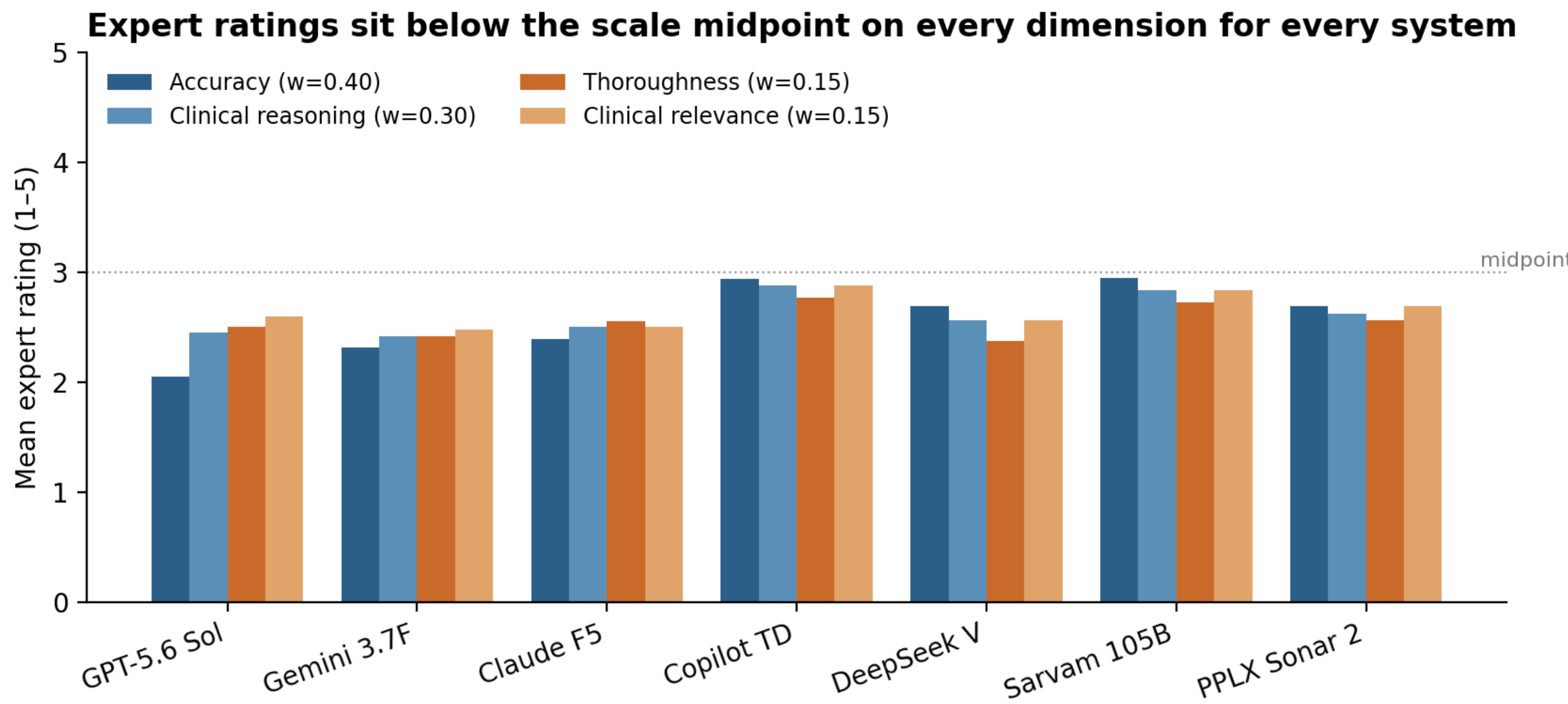

**Figure 2.** Mean expert rating (1–5) per rubric dimension per system over itemised free-response grades. No system reaches the midpoint on any dimension.

The quality axis and the verdict axis measure different things, and the data shows it. DeepSeek Vision posts a mean accuracy rating of 2.69, tied for third highest, while tying for the lowest resolution rate: when it reads the film correctly it reasons adequately, but it fails the cases that matter. ChatGPT 5.6 Sol shows the mirror pattern: the lowest mean accuracy rating (2.05) alongside mid-band reasoning and relevance scores, fluent clinical prose built on misread films. A benchmark reporting either axis alone would tell a materially different, and materially incomplete, story about both systems; this is the multimodal counterpart of the correct-verdict/defective-process dissociation documented for text-based clinical agents [6, 12].

## 4.3 Format-dependent competence: the same case, a different verdict

The six paired scenarios yield 42 matched model×scenario judgments in which the patient, the image, and the clinical question are identical and only the response format differs. Sixteen of the 42 verdicts (38.1%; 95% CI 25 to 53%) flip between formats: 10 from fail-as-MCQ to pass-as-free-response, and 6 in the opposite direction (Figure 3, Table 4). The interval excludes zero decisively, the format effect is real, while the directional split is statistically balanced (exact sign test, $p = 0.45$): neither format is systematically easier. Each has a distinct failure signature.

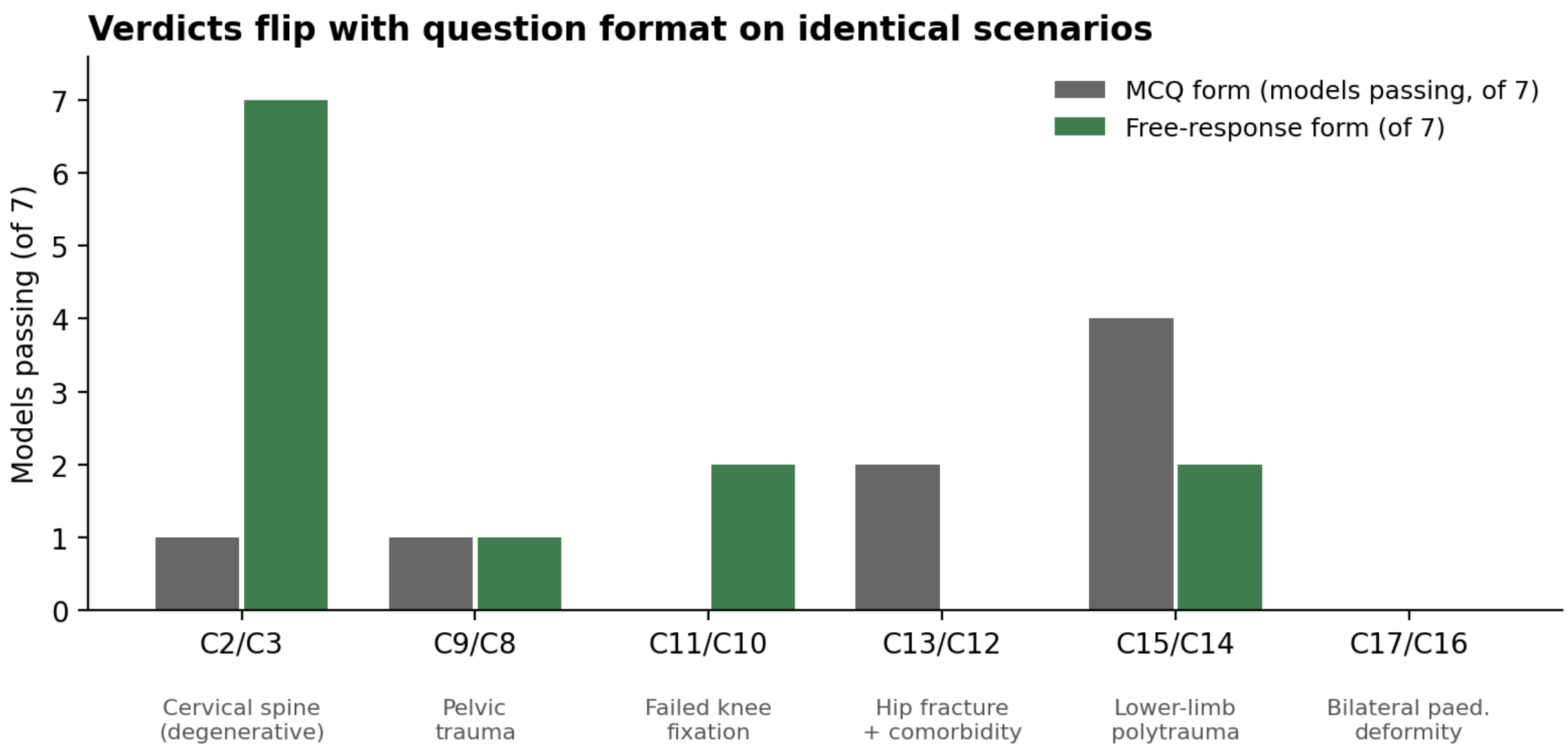

**Figure 3.** Models passing (of 7) on each paired scenario in its MCQ form (grey) and free-response form (green). The direction of the gap reverses across scenarios: format is not a difficulty dial but a different measurement.

**Table 4. Paired-format analysis.** Systems passing each form of the six paired scenarios, and the number of verdict flips in each direction. Held-out scenarios are identified by clinical domain only.

| Scenario (MCQ / free-response case) | Pass, MCQ | Pass, free-resp. | Flip: MCQ-only pass | Flip: free-resp.-only pass |
|---|---|---|---|---|
| Cervical spine, degenerative (C2 / C3) | 1 / 7 | 7 / 7 | 0 | 6 |
| Pelvic trauma, X-ray + CT (C9 / C8) - released | 1 / 7 | 1 / 7 | 1 | 1 |
| Failed knee fixation (C11 / C10) | 0 / 7 | 2 / 7 | 0 | 2 |
| Hip fracture with comorbidity (C13 / C12) | 2 / 7 | 0 / 7 | 2 | 0 |
| Lower-limb polytrauma (C15 / C14) - C14 released | 4 / 7 | 2 / 7 | 3 | 1 |
| Bilateral paediatric deformity (C17 / C16) | 0 / 7 | 0 / 7 | 0 | 0 |
| **Total (42 judgments)** | **8/ 42** | **12/ 42** | **6** | **10** |

## 4.4 The systems fail together, and often identically

Eleven of the 32 cases (34.4%; 95% CI 20 to 52%) were failed by all seven systems; 20 of 32 (62.5%) were passed by at most one system, and exactly one case was passed by all seven (Figure 4 and Figure 5). Where every system failed, they usually failed the same way: on one held-out hip film, all seven missed the same occult second fracture; on a held-out revision film, all seven missed the same subtle chronic finding that changes the whole strategy; on the bilateral deformity cases, no system graded the two sides separately. Two analyses sharpen the point. Agreement: across the 21 pairs of systems, average agreement on verdicts is 74.4% (range 66 to 88%), against 66.8% expected if failures were independent, so they cluster beyond chance. Convergence: on 9 of the 11 multiple-choice cases a majority of systems chose the same wrong option, and on 7 of 11 at least five of seven did. Six of seven chose the identical surgical option for a patient needing conservative care; six of seven ordered the same serious-disease work-up on the benign text-only control; five of seven chose the same overly extensive operation on the released pelvic case. When independently built systems miss the same finding and agree on the same wrong plan, asking a second vendor is not really a second opinion.

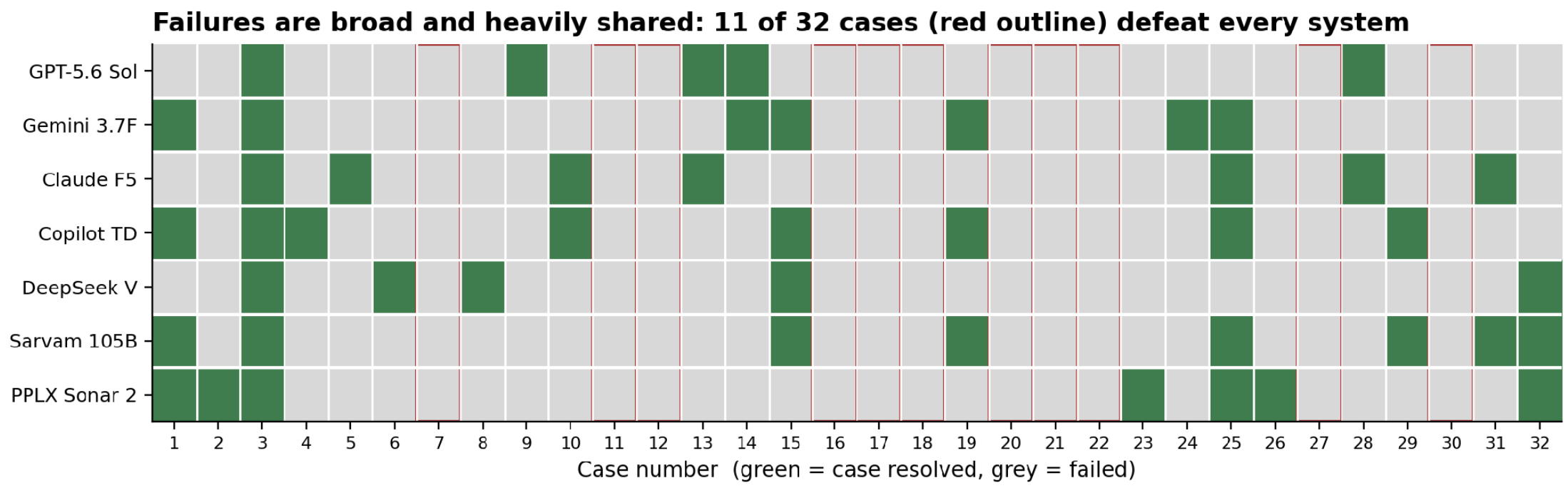

**Figure 4.** Pass and fail across all 32 cases and 7 systems (green = passed; red outline = the 11 cases every system failed). Failures cluster by case, not by system: difficulty is a property of the clinical material.

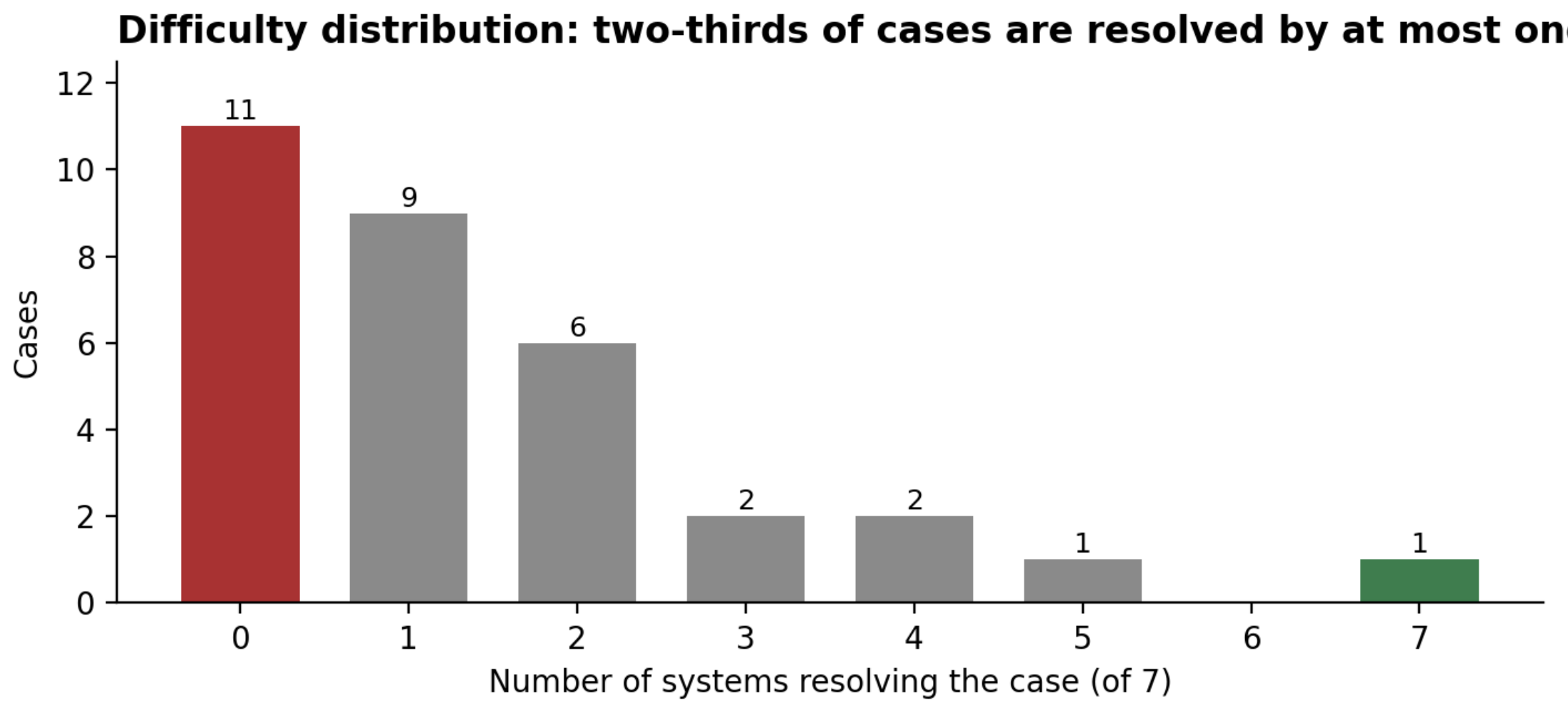

**Figure 5.** Distribution of cases by the number of systems resolving them. 20 of 32 cases are resolved by at most one system; a single case is resolved by all seven.

Four consistency observations from the grading log complete the picture. The adversarial-history control gives a process-versus-outcome data point inside our own multiple-choice set: on Case 1, all seven systems chose the correct option, yet three still failed because their image interpretation was wrong; option accuracy reads 100%, the process-level pass rate 57%. Systems also disagreed on identical inputs even when all were wrong; on one held-out revision case, one proposed a two-stage operation and another a single-stage, both missing the finding that decides between them. The same case put to a system on different dates sometimes returned different answers. And the failing answers arrived fluent and confident: no system said it was unsure, asked for better imaging, or declined to answer on any case it went on to fail, a missing-abstention pattern reported elsewhere as well [12].

## 5 The Failure Catalogue: 18 Recurring Discrepancies in Five Families

The lead grader wrote an analysis for every failing response and a commentary on notable cases, and the review panel checked these against the saved conversations. Coding those notes yields 18 distinct failure modes, each seen in multiple evaluations (Table 5). They sort into five families, ordered by where in the clinical pipeline the failure happens: analysing the image (F1), reasoning from findings to a plan (F2), integrating the image with the clinical report and history (F3), doing all of this consistently (F4), and doing it regardless of how the question is asked (F5). F1 and F3 are the two input channels every case exercises, and failures show up in each channel on its own as well as where they meet: the report-analysis failures occur even on the text-only control, while in the notes a wrong plan usually traces back to a misread or omission two steps earlier, which is why accuracy carries the largest rubric weight. Held-out discrepancies are described at pattern level.

**Table 5. The 18 failure modes observed and logged by the clinical team.** Families: F1 image analysis and grounding; F2 diagnostic and management reasoning; F3 clinical-report and context integration; F4 consistency and reliability; F5 question-format effects. † = released in full in Appendix A.

| # | Discrepancy | Family | Representative evidence |
|---|---|---|---|
| D1 | Temporal inconsistency: the same case, put to the same system on different dates, returns materially different answers | F4 | C18†, C15, C28 |
| D2 | Cross-system inconsistency: conflicting diagnoses or plans on identical input, including conflicting surgical strategies built on the same misread | F4 | C29, C18† |
| D3 | Wide variability in accuracy, reasoning, thoroughness, and relevance, even among answers graded correct | F4 | Grading log |
| D4 | Gross misreading of the X-ray, CT, or MRI: the film is described as showing an injury pattern it does not show | F1 | C12/C13, C20, C22, C30 |
| D5 | Incomplete image description: conspicuous findings are reported while decisive findings are left out (for example, no wall or column characterisation of a pelvic injury) | F1 | C8†, C29 |
| D6 | Wrong diagnosis and wrong management together: the misread carries through from start to finish | F2 | C8†, C7 |
| D7 | Wrong diagnosis: the primary condition is misidentified, so the management question never truly begins | F2 | C18†, C19, C25, C7 |
| D8 | Wrong management on a correct or partial diagnosis, including surgery that is too extensive and systemic treatment where local treatment is enough | F2 | C9†, C18†, C11, C21 |
| D9 | Highly variable depth of explanation for the imaging and the plan, from exhaustive narration to bare assertion | F4 | Grading log |
| D10 | A history that conflicts with the imaging goes undetected: the system follows the story and bends the film to fit it | F3 | C1 |
| D11 | The verdict on the identical scenario changes between multiple-choice and free-response form, in both directions | F5 | C8/C9†, C14†/C15 |
| D12 | Implants and hardware present in the patient are not identified; in some responses, hardware that was never placed is invented | F1 | C7, C24, C29 |
| D13 | No commitment to a definitive implant or procedure where surgery is indicated: plans stay generic (“lump-sum”) rather than exact | F2 | C4, C10 |
| D14 | Bilateral disease is not graded side by side: one treatment is prescribed for two sides of different severity | F2 | C16/C17, C25 |
| D15 | The clinical scenario is not integrated with the presentation, including escalation of a benign presentation whose explanation is in the vignette | F3 | C6, C21 |
| D16 | Hairline and subtle fractures visible on the film are missed | F1 | C22, C23, C26, C32 |
| D17 | Poor judgment between conservative and surgical management, with a consistent tilt toward intervention | F2 | C2/C3, C5, C19, C27, C28 |
| D18 | Multiple injuries in a single film are missed: the first finding ends the search | F1 | C14†, C20, C22, C31 |

**Image analysis (F1).** The most dangerous image failures were partial reads. A system would find one abnormality, report it confidently, and stop looking. On the released Case 14, a film with two fractures, only two systems identified both; one found the femur fracture and missed the patella, and three failed to make a correct diagnosis at all. On a held-out hip film, all seven missed a hidden second fracture; the case commentary calls the omission a blunder, and every plan downstream was built on an incomplete injury map. On the released pelvic case (C8), the recurring grading note makes the problem concrete: the injury was mentioned, but nobody characterised the specific wall and column involvement that fixation planning needs. Hardware showed the same weakness. In one held-out post-operative case, systems invented a fracture and a fixation procedure that were never there, while missing the hardware actually in the patient. In two other cases they could not name the implant that decides the revision plan, and in one of those, all seven also missed a subtle chronic finding pointing to an old infection. The released Case 9 adds a plain ingestion failure: one system stated in its explanation that the provided 3D CT scan had not been uploaded. These are exactly the misses that cause real-world harm, and independent fracture-detection studies of the same consumer systems report the same gap [16, 17].

**Diagnostic and management reasoning (F2).** The clearest public example is the released Case 18: a 73-year-old woman with knee pain and swelling after a fall, three years after a total knee replacement. All seven systems failed. Most diagnosed a periprosthetic fracture but could not agree where it was, placing it in the femur or the tibia from the same single X-ray, and two went on to recommend a revision knee replacement, which the grading notes describe as not at all needed. The same pattern runs through the held-out cases: plans are lump-sum rather than exact, the guideline gets reproduced, and the film goes unread. Side-specific judgment failed the same way, with one treatment prescribed for two differently graded sides on the deformity cases. And there is a consistent tilt toward doing more: systemic treatment where local treatment was enough, surgery where conservative care was correct, and, on the released Case 9, the overly extensive combined approach chosen by five of seven systems. Wherever the option space contains a more aggressive plan, the current systems drift toward it.

**Clinical-report analysis (F3).** On Case 1, three of seven systems read the film through the false story or gave up on independent image analysis altogether. The grading commentary draws the practical conclusion: when a non-medical user enters inaccurate history next to correct imaging, the chance of a false result rises sharply. On Case 6, six of seven escalated a benign presentation toward the same serious-disease work-up. The systems lean on the narrative too hard in both directions.

**Consistency and reliability (F4).** A clinician who sees the same film twice should reach the same conclusion twice. The systems often did not. The released Case 18 shows the problem at both levels. Within one system, ChatGPT gave a different diagnosis on an earlier date than on the date it was graded. Across systems, the seven placed the supposed injury in different bones, some in the femur and some in the tibia, from the same single X-ray. The held-out cases repeat the pattern: one system chose different options for the same multiple-choice case on different dates, and on a revision case two systems proposed opposite surgical strategies while both missed the finding that decides between them. Even correct answers varied widely in quality. On the released Case 14, the two passing responses scored 4.55 and 3.45 out of 5, a gap that a pass rate alone would hide. Depth of explanation varied just as much, from detailed walk-throughs of the film to bare one-line assertions. For a user, this means a single answer from a single system is a sample, not a settled opinion; asking again, or asking another system, can return a different verdict on the same patient.

**Question-format effects (F5).** Six scenarios appear twice in the benchmark, once as a multiple-choice question and once as free response, with the same patient and the same imaging. In 16 of the 42 matched pairs of verdicts, the same system passed one form and failed the other, and the flips ran in both directions. The grading notes point to two different mechanisms. In free response, a system can pass by describing every reasonable option somewhere in its answer; forced to pick one, the hedging is exposed. In multiple choice, the option list can do part of the work: on the polytrauma pair, only two systems produced an acceptable free-response answer on the released Case 14, with the rest missing the second fracture or failing to diagnose the injury at all, while four picked the right option on its multiple-choice twin, and the grader notes that the options helped them reach a better diagnosis. The released pelvic pair shows both directions at once: ChatGPT passed only the multiple-choice form, while DeepSeek passed only the free-response form and, in the multiple-choice form, said the provided 3D CT had not been uploaded. A benchmark that uses only one format would miss half of these failures.

**What the catalogue suggests doing.** The five families are not equally hard to fix. F5 is an evaluation-design issue and fixable now, by testing in both question formats. F4 becomes measurable with repeated runs. F2 and F3 look like data problems, since the corrective signal exists only in expert judgments about individual hard cases. F1 runs deepest. Current vision training seems to reward finding a finding rather than accounting for the whole film, so we propose secondary-finding recall, the fraction of expert-identified findings per image that a system reports, as a metric future multimodal medical evaluations should adopt; ClinEdge's annotations already support computing it.

## 6 Discussion, Scope, and Limitations

**What the results mean.** ClinEdge isolates the step that record-based and conversation-based benchmarks leave outside the loop: look at the primary evidence under imperfect context, then commit to a plan. That step is where these systems break: not on knowledge, which they clearly have, but on seeing everything in the film, trusting the evidence over the story, and committing to the right plan for the specific patient. For deployment, the findings stack badly: the systems answered every case fluently and confidently, including the third of cases all of them got wrong; they lean toward over-treatment; they follow a wrong history over a correct film; and they fail identically, so cross-checking vendors gives little protection. For benchmark builders, three practices follow: report uncertainty rather than bare scores; grade the reasoning and the image reading, not just the final answer (on our adversarial control, option accuracy reads 100% while the process-level pass rate is 57%); and keep hard cases held out, never reading edge-case suites as prevalence estimates. For model training, the signal needed to fix most failures here does not exist in guideline text or exam banks. It exists only in expert judgments about individual hard cases: which finding was hidden, why the default operation was wrong for this patient, what the hardware was. Expert-annotated real cases of that kind look like the scarcest and most valuable data for closing these gaps, and Appendix A shows what one unit of it looks like.

**Limitations.** ClinEdge is a small, deliberately difficult suite of 32 cases, so the results describe how systems handle hard real-world cases, not their average performance, and the failure rates should never be read as everyday error rates. This first release reports each system's single response per case and does not yet include a physician comparison arm; both a physician baseline and a repeated-run protocol are planned for the next version. The current cases lean toward musculoskeletal medicine, and expansion to other specialties is already under way.

## 7 Release, Access, and Future Scope

Public with this paper: the four cases of Appendix A in full; the complete evaluation framework (rubric, weights, anchors, the pass or fail criterion, and the protocol); the 18-mode catalogue; a domain-level roster of all 32 cases; and aggregate results, with every model conversation preserved as a replayable link. Held out: the other 28 cases, including all questions, imaging, options, and answer keys. Frontier AI labs and developers of healthcare-focused AI systems interested in evaluating their models against the held-out ClinEdge set are invited to contact the authors at research@futrbridge.com.

**Future scope.** We plan to grow the benchmark along four lines. First, more departments: new specialties such as cardiology, oncology, dermatology, and emergency medicine, and deeper coverage of those represented by only one to three cases today, such as paediatrics, ophthalmology, and internal medicine, each with its native imaging and report material (ECG, echocardiography, neuroimaging, histopathology, dermoscopy, laboratory panels). Second, more diverse patients and settings: a broader mix of age, sex, comorbidity, and geography across multiple institutions, including low-resource settings where consumer AI use is highest and imaging quality varies most. Third, more cases per failure family, enough for per-family scores. Fourth, richer case forms: longitudinal cases with follow-up imaging, and cases where the correct answer is to abstain or escalate.

## Authors, Ethics, and Disclosures

**Contributions.** Dr. Anurag Sharma and Aditya Sharma conceived the project and defined the vision and direction of the benchmark. Dr. Anurag Sharma, MS - Orthopaedics, served as lead clinical expert: he curated the cases, wrote every scenario and option set, designed the rubric, performed the primary grading of all 224 outputs against each patient's documented course, and wrote the failure notes that ground Section 5. Aditya Sharma, AI Engineer, led the AI engineering: the evaluation framework and protocol, all model evaluations and logging, the analysis pipeline, statistics, figures, and the manuscript. Sapna Baalwan, Project Manager, led project management and coordination across the AI and healthcare workstreams and, drawing on her healthcare background, helped ensure the clinical work followed applicable healthcare guidelines. The clinical review panel reviewed the cases and grades as described in Section 3.4. The authors approved the manuscript.

**Clinical review panel.** The following physicians served on the ClinEdge Clinical Review Panel, independently reviewing the cases for clinical validity and the grades against the saved conversations: Dr. R. P. Asat (MS Orthopaedics, SNMC Jodhpur), Senior Professor and Head of the Department of Orthopaedics, SMS Medical College & Hospital, Jaipur (SMS MC&H); Dr. Pankaj Jain (MS Orthopaedics, MAMC Delhi), Senior Professor and Head of Unit, SMS MC&H; Dr. Devkant Meena (MS Orthopaedics, AIIMS Delhi), Head of Unit, SMS MC&H; Dr. Anurag Dhaker (MS Orthopaedics, SNMC Jodhpur), Senior Professor and Head of Unit, SMS MC&H; Dr. Prashant Modi (MS Orthopaedics, UCMS Delhi), Professor and Head of Unit, SMS MC&H; Dr. Arun K. Sharma (MS Orthopaedics, PGI Chandigarh), Professor and Head of Unit, SMS MC&H; Dr. S. K. Goyal (MS Orthopaedics, RNT Medical College Udaipur), Head of Unit, SMS MC&H; Dr. Ramdayal Sahu (MS Urology), Assistant Professor, SMS MC&H; Dr. Anil K. Sharma (MS Orthopaedics, SMS Medical College); Dr. Avatar Singh Balavat (MS Orthopaedics, SPMC Bikaner), Associate Professor, SMS MC&H; Dr. Shivprakash Sharma (MD, Preventive and Social Medicine, RNT Medical College Udaipur), Associate Professor, RUHS Medical College and Hospital; Dr. Vishu Pratap Singh Chauhan (MS Orthopaedics, RNT Medical College Udaipur), Assistant Professor, RNT Medical College and Hospital, Udaipur; Dr. Lokpal Singh Bhati (MS Orthopaedics, SMS MC&H), Assistant Professor, Jaisalmer Medical College and Hospital; Dr. Manish Mehta (MS Orthopaedics, SMS MC&H), Assistant Professor, Tonk Medical College and Hospital; Dr. Laxman Chaudhary (MS Orthopaedics, SMS Medical College), Assistant Professor, AIIMS Jodhpur; Dr. Vineet Kuntal (MS Orthopaedics, SMS MC&H), Assistant Professor, SMS MC&H; Dr. Tushar Sharma (MS Orthopaedics, NSCB Medical College Jabalpur), Assistant Professor, SMS MC&H; Dr. Kaushalesh Sharma (MS Orthopaedics, SMS MC&H); Dr. Mahesh Gurjar (MD, Physical Medicine and Rehabilitation, SMS MC&H).

Institutional positions are listed to identify panel members' clinical roles only; the institutions named did not sponsor, fund, or participate in this study, and the views expressed are those of the authors.

**Acknowledgements.** The authors thank Dr. Harendra Singh Gurjar (BPT, Jaipur) and Dhruv Sharma (BPT candidate, MJRP Jaipur) for research assistance, and Divya Tarwadi and Lata Somai for their assistance with running the model evaluations, document preparation, data cleaning, and review.

**Patient data and ethics.** All cases come from routine clinical care and were fully de-identified before use: every direct and indirect identifier was removed, including names, contact details, record numbers, dates of care, institution names, and any markings on the imaging, and no information that could reveal a patient's identity appears in the cases, the grading records, or this paper. This is a retrospective analysis of anonymised records and imaging in which no patient was contacted and no care was altered; the evaluated systems received only the de-identified material.

**Competing interests and funding.** The authors curate and administer ClinEdge and may offer graded evaluation against its held-out set as a service to AI developers. No evaluated vendor funded, reviewed, or influenced this work, and the authors have no financial relationship with any evaluated vendor. This research received no external funding.

# Appendix A The Four Released Cases

These four cases are released in full: vignette, imaging, options and correct answer for the multiple-choice case, per-system verdicts and scores. They cover both question formats and two imaging configurations. In the tables, A = accuracy, R = clinical reasoning, T = thoroughness, C = clinical relevance, WS = weighted score.

### A.1 Case 8 (free-response)

**Vignette:** A 33-year-old man was involved in a road traffic accident five days ago and sustained a pelvic injury. What should be the next step in his management?

**Department:** Radiology, Orthopaedics, Implantology

**Image Type:** X-ray and CT Scan (Find image here)

**Table A1.** Case 8, per-system results

| System | Verdict | WS | A | R | T | C |
|---|---|---|---|---|---|---|
| DeepSeek Vision | **Pass** | 3.70 | 4 | 4 | 2 | 4 |
| ChatGPT 5.6 Sol | Fail | 1.60 | 1 | 2 | 2 | 2 |
| Gemini 3.7 Flash | Fail | 1.60 | 1 | 2 | 2 | 2 |
| Claude Fable 5 | Fail | 1.60 | 1 | 2 | 2 | 2 |
| Copilot (Think Deeper) | Fail | 1.60 | 1 | 2 | 2 | 2 |
| Sarvam 105B | Fail | 1.60 | 1 | 2 | 2 | 2 |
| Perplexity Sonar 2 | Fail | 1.60 | 1 | 2 | 2 | 2 |

### A.2 Case 9 (multiple-choice question)

**Vignette:** A 33-year-old man was involved in a road traffic accident five days ago and sustained a pelvic injury. What should be the next step in his management?

**Options:**
1. Fixation with anterior and posterior approach
2. Fixation with anterior approach only
3. Fixation with posterior approach only

Correct Answer is Option 2 - Fixation with anterior approach only

**Department:** Radiology, Orthopaedics, Implantology

**Image Type:** X-ray and CT Scan (Find images here)

**Table A2.** Case 9, per-system results

| System | Outcome | Verdict |
|---|---|---|
| DeepSeek Vision | Option 1 | Fail |
| ChatGPT 5.6 Sol | **Option 2** | **Pass** |
| Gemini 3.7 Flash | Option 1 | Fail |
| Claude Fable 5 | Option 3 | Fail |
| Copilot (Think Deeper) | Option 1 | Fail |
| Sarvam 105B | Option 1 | Fail |
| Perplexity Sonar 2 | Option 1 | Fail |

### A.3 Case 14 (free-response)

**Vignette:** He is 25 year old patient who had a RTA 2 days back. What is the diagnosis?

**Department:** Radiology, Orthopaedics

**Image Type:** X-ray (Find image here)

**Table A3.** Case 14, per-system results

| System | Verdict | WS | A | R | T | C |
|---|---|---|---|---|---|---|
| DeepSeek Vision | Fail | 1.00 | 1 | 1 | 1 | 1 |
| ChatGPT 5.6 Sol | Pass | 3.45 | 3 | 4 | 3 | 4 |
| Gemini 3.7 Flash | Pass | 4.55 | 5 | 4 | 4 | 5 |
| Claude Fable 5 | Fail | 2.60 | 2 | 3 | 3 | 3 |
| Copilot (Think Deeper) | Fail | 1.00 | 1 | 1 | 1 | 1 |
| Sarvam 105B | Fail | 1.00 | 1 | 1 | 1 | 1 |
| Perplexity Sonar 2 | Not Evaluated - technical failure* | | | | | |

*Perplexity was not evaluated for this case because the X-ray image failed to upload.

### A.4 Case 18 (free-response)

**Vignette:** A 73-year-old woman presented with pain and swelling in her knee after falling at home the previous evening. She underwent total knee replacement (TKR) three years ago and had been walking without difficulty since then. The patient's X-ray is shown. What is the definitive diagnosis, and what are the best treatment options for her?

**Department:** Radiology, Orthopaedics

**Image Type:** X-ray (Find image here)

**Table A4.** Case 18, per-system results

| System | Verdict | WS | A | R | T | C |
|---|---|---|---|---|---|---|
| DeepSeek Vision | Fail | 2.00 | 2 | 2 | 2 | 2 |
| ChatGPT 5.6 Sol | Fail | 2.75 | 2 | 3 | 3 | 4 |
| Gemini 3.7 Flash | Fail | 1.00 | 1 | 1 | 1 | 1 |
| Claude Fable 5 | Fail | 1.00 | 1 | 1 | 1 | 1 |
| Copilot (Think Deeper) | Fail | 2.00 | 2 | 2 | 2 | 2 |
| Sarvam 105B | Fail | 1.00 | 1 | 1 | 1 | 1 |
| Perplexity Sonar 2 | Fail | 1.00 | 1 | 1 | 1 | 1 |